\documentclass[conference]{IEEEtran}
\IEEEoverridecommandlockouts
\usepackage{cite}
\usepackage{amsmath,amssymb,amsfonts}
\usepackage{algorithmic}
\usepackage{graphicx}
\usepackage{textcomp}
\usepackage{xcolor}
\usepackage{hyperref}
\usepackage{url}
\hypersetup{
colorlinks=true,
urlcolor=cyan
}

\def\BibTeX{{\rm B\kern-.05em{\sc i\kern-.025em b}\kern-.08em
    T\kern-.1667em\lower.7ex\hbox{E}\kern-.125emX}}

\begin{document}

\title{Harnessing expressive capacity of Machine Learning modeling to represent complex coupling of Earth's auroral space weather regimes  \\
\thanks{With support from DARPA under Department of the Interior Heliosphere to Earth's Atmosphere Rendering via Excellent Artificial Intelligence Training (HEARTBEAT) Award D19AC00009 to Georgia Institute of Technology with subaward to ASTRA as well as acknowledging R. McGranaghan's NASA Early Career Investigator Program grant \#80NSSC21K0622.}
}

\author{
\IEEEauthorblockN{Jack Ziegler}
\IEEEauthorblockA{\textit{Atmospheric Space and Technology Research Associates (ASTRA)} \\
Louisville, CO, USA \\
jziegler@astraspace.net}
\and
\IEEEauthorblockN{Ryan M. Mcgranaghan}
\IEEEauthorblockA{\textit{ASTRA} \\
Louisville, CO, USA \\
rmcgranaghan@astraspace.net}
}

\maketitle
\begin{abstract}
We develop multiple Deep Learning (DL) models that advance the state-of-the-art predictions of the global auroral particle precipitation. We use observations from low Earth orbiting spacecraft of the electron energy flux to develop a model that improves global nowcasts (predictions at the time of observation) of the accelerated particles.  Multiple Machine Learning (ML) modeling approaches are compared, including a novel multi-task model, models with tail- and distribution-based loss functions, and a spatio-temporally sparse 2D-convolutional model.  
We detail the data preparation process as well as the model development that will be illustrative for many similar time series global regression problems in space weather and across domains. Our ML improvements are three-fold: 1) loss  function  engineering; 2) multi-task learning; and 3) transforming the task from time series prediction to spatio-temporal prediction. Notably, the ML models improve prediction of the extreme events, historically obstinate to accurate specification and indicate that increased expressive capacity provided by ML innovation can address grand challenges in the science of space weather. 
\end{abstract}


\section{Introduction}
`Space weather' refers to the tangible effects that solar energy has on our technologies, space- and ground-based infrastructure, and our daily lives \cite{Schrijver_2015}. It is a wide-reaching science that spans from the sun to the Earth's surface, requiring the integration and sophisticated representation of heterogeneous and complex data. A key to understanding the weather in outer space is how regions between the Sun and the Earth's surface are connected. One challenging space weather phenomena to model is the coupling of energy from the solar wind into the magnetosphere and down into the Earth's upper atmosphere (100–1,000 km altitude). Particles moving along magnetic field lines `precipitate' into this region, carrying energy and momentum which drive space weather and a host of societal impacts such as satellite damage or destruction and disruptions to communication systems. 

Predicting auroral particle precipitation is a grand challenge in space weather because it is a complex result of driving parameters from the sun, solar wind, and Earth space environment. We need prediction capabilities for the global spatio-temporal patterns of precipitation, a problem that has been obstinate to empirical approaches. In this work we advance the state of precipitation prediction by applying cutting-edge techniques from the field of ML. 

The specific problem formulation is to use observations of the solar wind (the carrier of solar energy throughout interplanetary space), the magnetosphere, and the Earth's space environment to predict the spatio-temporal pattern of total electron energy flux as observed by the Defense Meteorological Satellite Program (DMSP) satellites. These fluxes represent electrons that come from the magnetosphere and move along magnetic field lines into the upper atmosphere (100-1000 km altitude). When they impact molecules (e.g., our atmosphere) they cause ionizations and visible emissions -- the aurora.

\begin{figure}[htbp]
\centerline{\includegraphics[width=.85\linewidth]{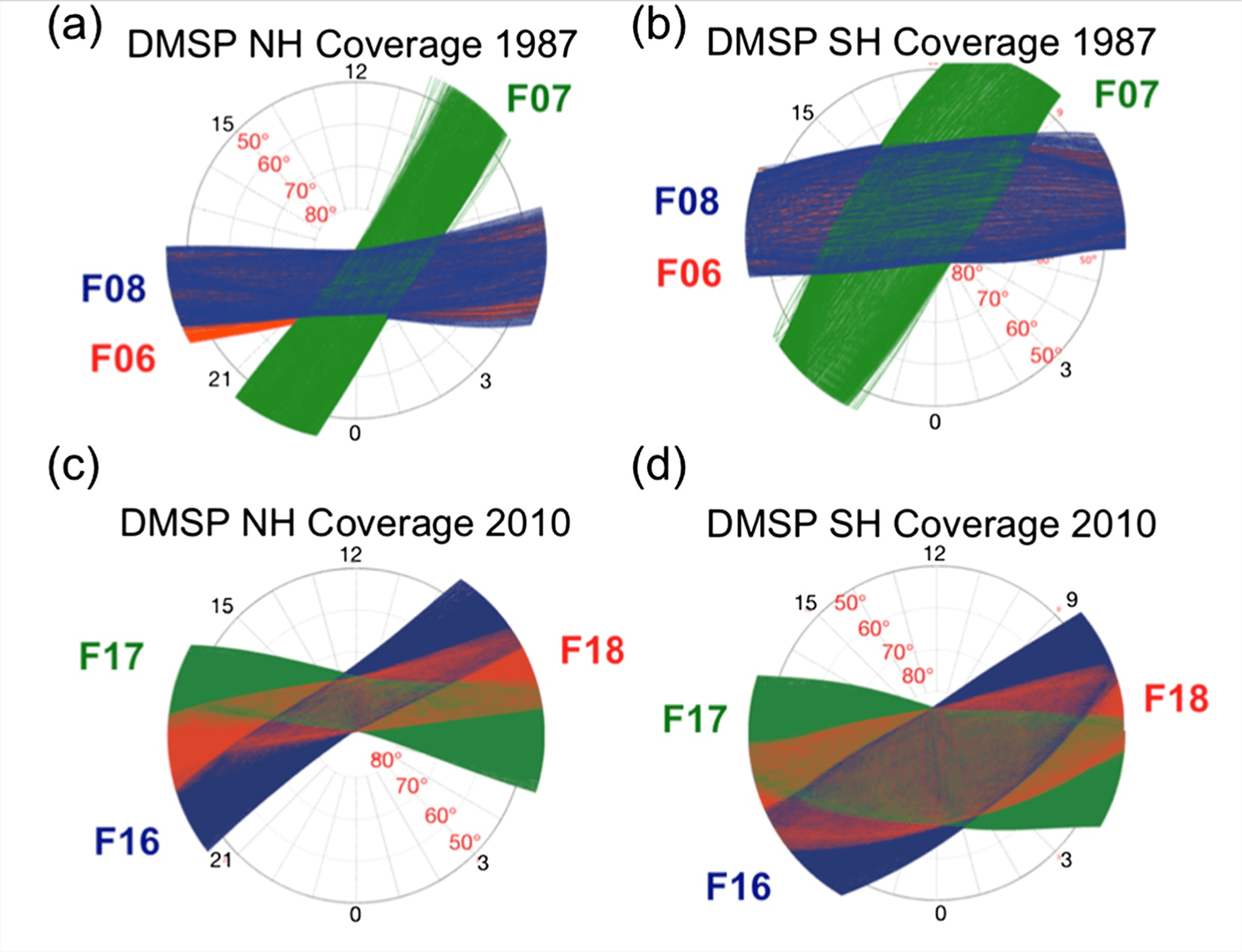}}
\caption{Combining NH and SH observations from multiple satellites to create a dataset that nearly covers all local time (0 to 24 hour time zones) \cite{Mcgranaghan_2016}.}
\label{coverage}
\end{figure}

Auroral particle precipitation is a critical input to global circulation models (e.g., \cite{Scherliess_2019}), which describes the effects that the energy deposited by electrons and ions has.  This causes ionizations that change the ionosphere and affect radio communications, heating the thermosphere leading to enhanced neutral density. This changes the orbits of all objects within the upper atmosphere and the coupling to the entire Earth system.
Our models use DMSP data dating back to 1987. DMSP spacecraft carry a particle detector (SSJ) which is sensitive to particles with characteristic energies between 30 eV and 30 keV.  Figure \ref{coverage} shows polar plots that each have three satellites, illustrating their coverages over different years \cite{McGranaghan_2015}. 

In this work, we first treat each satellite measurement independently and then use all measurements available during a small window as simultaneous to construct multi-spatial observations at a given time step.
Our models achieve $>50\%$ reduction in errors from the current state-of-the-art model, The Oval Variation, Assessment, Tracking, Intensity, and Online Nowcasting Prime (OVATION Prime) model \cite{Newell_2009, Machol_2012}. These new models better capture the dynamic changes of the auroral electron energy flux, and provide evidence that it can capably reconstruct mesoscale phenomena. 

The key contributions of this work are:
\begin{enumerate}
    \item Advancing the state-of-the-art ML applied to auroral particle precipitation; 
    \item Assessment of the efficacy of increasing ML model expressive capacity (loss function engineering, multi-task learning, and transforming time series prediction to spatio-temporal prediction);
    \item New lessons learned to advance ML research through application to novel science data. 
\end{enumerate}

Item 2) is of important significance to other ML modeling problems dealing with sparse 2D temporal data, a common problem for any model using satellite sensor data or data from any sensor that is following 1-D traces on a 2D domain.

\section{Related Work}

Existing models of particle precipitation do not adequately capture the mesoscale (100-400 km scale) spatial features and take advantage of only a small number of the important driving parameters. Typically, in the upper atmosphere, simple statistical and empirical patterns of the particle precipitation coarsely parameterized by small subsets of the available information are used to quantify the complex processes \cite{Newell_2015}. 

Particle precipitation models such as the widely used \cite{Hardy_1985} model are driven only by the planetary Kp index whose time resolution is three hours and prevents the model from capturing changes that occur on a shorter time-scales. More recently, models have embraced the need to incorporate more information and to utilize different sets of organizing parameters as a cornerstone to advancing our understanding and ability to predict particle precipitation. Lacking from the \cite{Hardy_1985} and related empirical models is a realistic representation of the temporal dependence between solar wind input and particle precipitation response. It has also become apparent that increasing the spatial resolution of the models reveals physically meaningful mesoscale behavior (defined in this work to be ~100–400 km, 1-3$^{\circ}$ magnetic latitude (MLAT) width).

The series of OVATION, later becoming OVATION Prime (OVP) models \cite{Newell_2002, Newell_2009, Newell_2010a, Newell_2010b, Newell_2010c} utilized solar wind parameters, making the implicit assumption that these parameters are the cause and the observed electron precipitation the effect. Moreover, data from geospace (e.g., geomagnetic indices) and from LEO satellites assumes that these parameters provide information of the simultaneous and global electron precipitation distribution (e.g., 'state descriptors' \cite{Gjerloev_2018}). OVP was a significant advance, however is limited by a reliance on relatively simple relationships to organizing parameters and spatial binning. The success and shortcomings of OVP give credence to the importance of organizing parameters and representation of driver-response relationship. \cite{McGranaghan_2021} took the first step to address these gaps and applied ML to create the current state-of-the-art model, PrecipNet. They showed that rich information exists in the spectrum of Heliophysics observational data and that careful representation of this information with ML can lead to profound progress. That work lays the foundation for the advance we present in this paper. 

\section{Advancing the Frontier of ML for Auroral Particle Precipitation} \label{methodology}

PrecipNet and related work represent the state-of-the-art for ML applied to auroral particle precipitation.  Against this backdrop, we raise the question, ``Does increasing the expressive capacity/complexity of the ML algorithm lead to demonstrable improvements in ML prediction for physical problems?'' We explore four new areas of the ML algorithm search space: 
\begin{enumerate}
    \item Multi-task learning models
    \item `Tail of the distribution'-specific custom loss functions
    \item Sample distribution weighted loss function
    \item Novel 2D, inverse convolutional\cite{McCann_2017} (Conv2D) model for sparse spatial-temporal training data
\end{enumerate} 
 In addition, we explore combinations of the four areas. 
 \cite{McGranaghan_2021} outlined a consistent and rigorous framework for assessing space weather models and we adopt their evaluation formalism (Section 6) to justify our quantifiable improvements.  

\subsection{Input Parameters}
Our models utilize the time-dependent NASA space weather data \url{https://omniweb.gsfc.nasa.gov/OMNIWeb} as inputs. The solar wind parameters and magnetic field values are primary examples. All of the details of our dataset and python scripts for data preparation are provided for reproducibility and extensibility \url{https://github.com/rmcgranaghan/HEARTBEAT} GIT repo. The focus of this work is on model improvements and comparisons and therefore the inputs and outputs are the same as in \cite{McGranaghan_2021}. Inputs are time series and the target to be predicted represents both 2D spatial and temporal variations. For any time step, only spatially sparse target observations exist (at only one or a few spatial locations from different simultaneous observing satellites). Thus, we must develop relationships between the input features and the particle precipitation response across many years to construct a dataset that is representative of both the spatial and temporal variations. 
\begin{itemize}
    \item \textbf{Inputs}: Global  solar wind observations and geomagnetic indices at five-min. cadence; \item \textbf{Target Output}: Total electron energy flux at any given location (nowcast), MLat and MLT (Magnetic Local Time (zone)), with (altitude variations ignored).
\end{itemize}
We pose the problem as regression, predicting the magnitude of the total electron energy flux using both instantaneous and time history observations. Note that a DMSP spacecraft observation is only available at one location at any given time. There may be multiple satellites providing observations at some but not all time steps. 

The solar wind input variables are collected by satellites orbiting between the Sun and the Earth and quantify the state of the solar energy that drives space weather. The geomagnetic indices are proxies for space weather activity based on magnetometer observations at Earth's surface and aggregated into global indices that quantify the near-Earth space environment (like a hurricane is described as e.g., category 1 - 5). 

The ML target is the total electron energy flux. Our withheld validation set consists of all DMSP satellite data from the F16 spacecraft in the year 2010, more than 50000 data points when using one-min. resolution. Taking a consecutive year of data for validation, rather than randomly sampling the full database, prevents influence by autocorrelations in the data.\\
\textbf{Model Inputs}: 148 features total (full list described in \cite{McGranaghan_2021}).
\begin{itemize}
    \item \textbf{Spatial and temporal inputs}: Magnetic Latitude (MLAT), Magnetic local time (MLT) (which can be seen in Figure \ref{coverage}: rings are increments of MLAT from the pole (90$^{\circ}$) and MLT is around the dial (the sun, 12 noon, is off to the top of the figures) 
\item \textbf{Global variables' time history inputs}:\\
 \textit{Instantaneous} (referenced to the target observation time): 0, -5, -10, and -15 min. \\
 \textit{Averaged}: -30 and -45 min. and -1, -3, -5, and -6 hrs.\\
 \textit{Geomagnetic indices}: AE, AL, AU, F107, SymH\\
 \textit{Solar wind-dependent and -derived parameters}:\\ Bx, By, Bz, Vsw, Psw, Vx, PC, Newell \cite{Newell_2007} and Borovsky \cite{Borovsky_2018} Coupling Functions
\end{itemize}
\textbf{Model output target}: Total Electron Energy Flux at time 0. Validation points are at points where satellites are but the models are easily evaluated at all spatial point at one time step for weather nowcasting.

By ``global'' inputs we mean that they represent a parameter for the whole Earth and do not have a particular spatial location, just a temporal variation. We explore more parsimonious input feature sets (less than the full 148 features), but those results are beyond the scope of this short paper. Also, note that the spatial inputs, MLAT and MLT, are in a reference frame that does not rotate with the earth.  The local time has the range of 0 to 24 hours and has 12 noon facing the sun.

\subsection{Data Challenges}
\begin{figure}[htbp]
\centerline{\includegraphics[width=.55\linewidth]{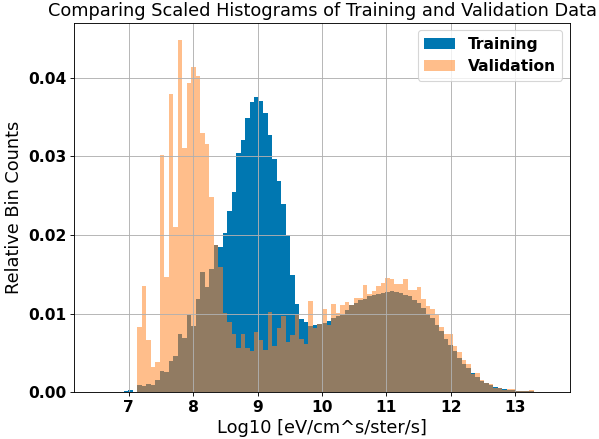}}
\caption{Showing scaled comparisons of training and validation data histograms. Note the smoothness of the training distribution due to more data points and the difference in the left peak or most common values of the target.}
\label{trainhisto}
\end{figure}

Auroral particle precipitation and space weather create numerous data challenges that advance to novel ML methods:
\begin{itemize}
    \item Target Transformation: The electron energy fluxes [eV/cm$^2$/s] vary over six orders of magnitude and require a log transformation. 
    \item Unrealistic fluxes: We remove target values that are deemed non-physical due to magnitude (justified by physical understanding of auroral particle precipitation). These were determined to be those exceeding the 99.995th percentile $(> 7.37 10^{13}\space eV/{cm}^2/sr/s)$.
    \item Imbalanced data: In space weather, like with terrestrial weather, the active periods are much less frequent than quiet times. Large fluxes tend to occur during storms, so our target variable exhibits a large right ``tail''.
    \item Multimodal target variable distribution: Target points composed of multiple ``modes'' of skewed Gaussian-like distributions (see Figure \ref{trainhisto}).
\end{itemize}
The one-sec. resolution target data is available but currently too large for training efficiently. Our PrecipNet-based models are able to train using the full one-min. resolution DMSP data in the order of 10 min. with a 12 GB Nvidia 2080Ti GPU.  The Conv2D model trains in a few hours. Preparing the combined dataset, cleaned OMNI (independent variables) and DMSP (dependent variables), is the most computationally inefficient process, demanding 128 GB of data to be downloaded and filtered to a lower 1-min Cadence, as well as the removal of outliers, and the calculation of moving averages of the input variables and derived quantities.   

When viewing a histogram of the logarithm of the target flux, Figure \ref{trainhisto}, multiple modes in the distribution are observed. There is evidence that these modes correspond to unique auroral spatial regions: sub-auroral (below the auroral oval), auroral, and in the polar cap (poleward of the auroral oval). The validation data distribution is also shown for reference, and its differing low flux mode (for the unique year 2010) poses a modeling challenge.

\subsection{Modeling Challenges}
This research responds to several modeling challenges:
\begin{itemize}
    \item Deconvolving spatial and temporal responses to inputs
    \item Extreme event predictions of rare space weather storms
    \item Evaluating 2D spatial performance of model
    \item Designing loss functions / metrics prioritizing large fluxes
    \item Predicting mesoscale variations and structures
\end{itemize}
 Our models need to deconvolve temporal and spatial responses as there is only one time-step for each 2D data sample's position in the orbit, at times there are no more than 3 satellites in orbit at once.  

\section{Model descriptions}
\subsection{Baseline Model}
\begin{figure}[htbp]
\centerline{\includegraphics[width=\linewidth]{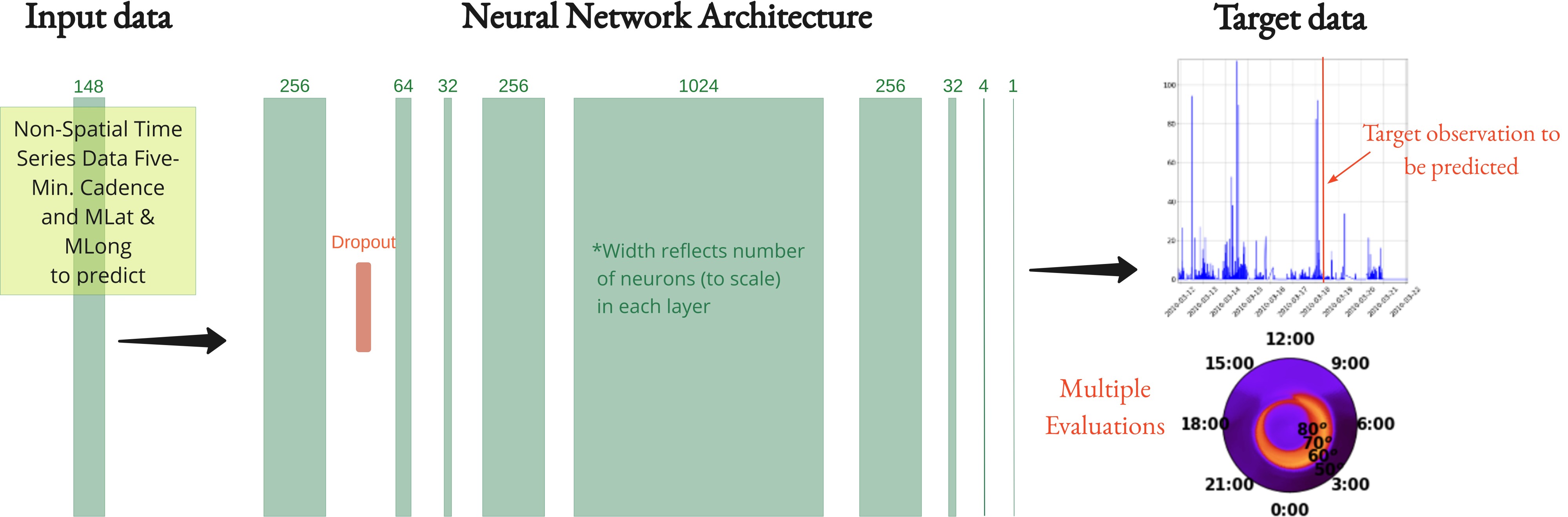}}
\caption{PrecipNet Baseline Model}
\label{baseline}
\end{figure}

We use PrecipNet\cite{McGranaghan_2021} as our baseline model against which to demonstrate improvement. 
The architecture of PrecipNet is detailed in Figure \ref{baseline}.
Each model we present builds from this baseline architecture. Three updates are tested: application of custom loss functions, use of a final layer for sequential classification and regression (multi-task model), and addition of a final 2D inverse/transposed convolutional layer (the `Conv2D' model).  This baseline model uses the mean square error (MSE) loss and therefore has been optimized to have the lowest MSE over all training data. However, using MSE on imbalanced data leads to sub-optimal accuracy for predicting rare high flux events.  Using the log 10 scale transformation helps with balancing the data but not enough.  

\subsection{Multitask Auroral Region Model}
\begin{figure}[htbp]
\centerline{\includegraphics[width=.85\linewidth]{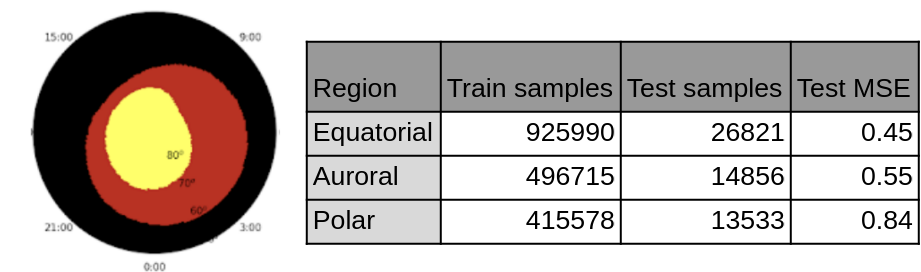}}
\caption{Auroral region classification model result for the northern hemisphere and top down view of the earth centered at the magnetic north pole from 45 to 90 degrees MLAT and the sun off to the top of the figure, showing the high activity 'red' auroral region, 'yellow' moderate activity polar region, and a 'black' low activity equatorial region.}
\label{regions}
\end{figure}

To separate the physically distinct processes that produce the multi-modal nature of the auroral flux distribution, we developed a 'multi-task' auroral boundary model that first classifies by auroral region and then regresses the flux value in each region. We chose a three-class classification task to separate sub-auroral, auroral, and polar preciptitation regions that are known to have distinct precipitation characteristics. The 2010-2014 subset (12 satellite years) of the DMSP data includes observations of the auroral boundary transition regions based on spatial gradients of the electron flux \cite{Kilcommons_2017}. We used these transitions to create a classification model of the auroral regions, one representative result is shown in Figure \ref{regions}.  The result of this region estimation model can be fed into the regression model for electron flux as an additional input or used to separate the data to train three models. It was also used as a classification result for training and validation model, as was the case of the final variation.

Using the auroral region classifier (90\% accurate classification over 45 to 90 degree MLat paths), as one experiment, the data was separated and three separate models were trained on each region. The table in Figure \ref{regions} show that the MSE for the polar region (yellow) is the largest and the (black) Equatorial the lowest.  This shows that the model is most accurate at predicting the black, low-activity sub-auroral region.  Surprisingly, the auroral region (red) (which contains the most dynamic precipitation) is only ~20\% higher in error while the polar region error is ~50\% higher. This means that the model is more accurate at representing the flux near the outer (lower MLat) auroral boundary than the inner.  This could also mean that it is possible that there is not enough input information to describe the polar region flux.

{\it Combining MSE and categorical cross-entropy}:
 Our final multi-task model predicts the auroral region type and the total electron energy flux simultaneously with a loss function that has two penalty terms: categorical cross entropy  and MSE. This model has two output branches, one side with three softmax values for predicting the class, and the other side which predicts the flux quantity for each region.  Note that in the loss function, only the model's predicted region value (one of the three regression values) is used in the MSE contribution. These models were inspired from the following blog post: \href{https://towardsdatascience.com/anchors-and-multi-bin-loss-for-multi-modal-target-regression-647ea1974617}{Anchors and multi-bin loss for multi-modal target regression}.
 \begin{equation}
 \begin{aligned}
  \textrm{Loss} =\; &\textrm{Mean}\big( (y_{\textrm{flux-true}}-y_{\textrm{flux-pred}})^2\big)\\
  & + \; \textrm{C.C.L.}\big(y_{\textrm{class-true}},y_{c\textrm{class-pred}}\big) \label{eq1}
 \end{aligned}
 \end{equation}
 where C.C.L. represents the categorical crossentropy loss (as defined by the Tensorflow Keras library).
 
\subsection{Custom Tail Loss Model}
In order to better capture the extreme values we adopt a similar approach to \cite{Qi52} where a logic-based function only activates in the tail regions of an unbalanced distribution of training data. The result is a ``tail-weighted loss'' function which is useful to capture extreme values of a distribution. 
\begin{equation}
\begin{aligned}
\textrm{Loss} =\; &{(y_{\textrm{true}}-y_{\textrm{pred}}})^2 (1 + a) \;\\ & \mathbf{\textrm{for}} \; y_{\textrm{true}} > y_r\; \mathbf{\&} \; y_{\textrm{pred}} < y_r \label{eq2}
\end{aligned}
\end{equation}
Our variant of this loss function in \cite{Qi52} was found to increase the accuracy in the tail region without significantly decreasing the accuracy in the lower flux region, demonstrated in Table \ref{error_percent_tail_loss}. For example a 26\% decrease in error is observed in the 90th precentile, while the accuracy at the upper half of the 50th percentile reveals no loss of accuracy.  This function is a sum of the MSE and a special term which is only $>$ 0 in particular circumstances where for a threshold, the $y_{\textrm{true}}$ value is above and $y_{\textrm{pred}}$ is below.  This additional term has the effect of penalizing the neural network when a true high value has a prediction below a predetermined threshold.  If the true value is also below this threshold or if both the true and predicted value are over the threshold, then there is no penalty.
This threshold $y_r$ is chosen, where $y$ is the target and $r$ signifies the ``right side'' of the tail. This value is within the tail region on the right side of the histogram.  When this condition is true, an additional loss penalty is added to the loss, proportional by the factor ``$a$'' to the MSE, $(y_{\textrm{true}}-y_{\textrm{pred}})^2$. 
In our Tail Loss results, empirical exploration led to the use of a sum of multiple $a$ and $y_r$ terms which we selected from points on the tail of the training distribution that were underrepresented by the original PrecipNet with just MSE loss.  The results in this work used five $(a,y_r)_i$ pairs of (2.5,12), (5,12.5), (10,13), (10,13.25), (10,13.5), where the units of $y_r$ are log10 [$eV/{cm}^2/ster/s$]. The values of '$a$' were found empirically and the target values where selected by seeing where on the right tail of the predicted training distribution (shown in Figure \ref{trainhisto}) the target range was insufficiently represented.  
\subsection{Distribution Re-balancing Loss Model}
\begin{figure}[htbp]
\centerline{\includegraphics[width=\linewidth]{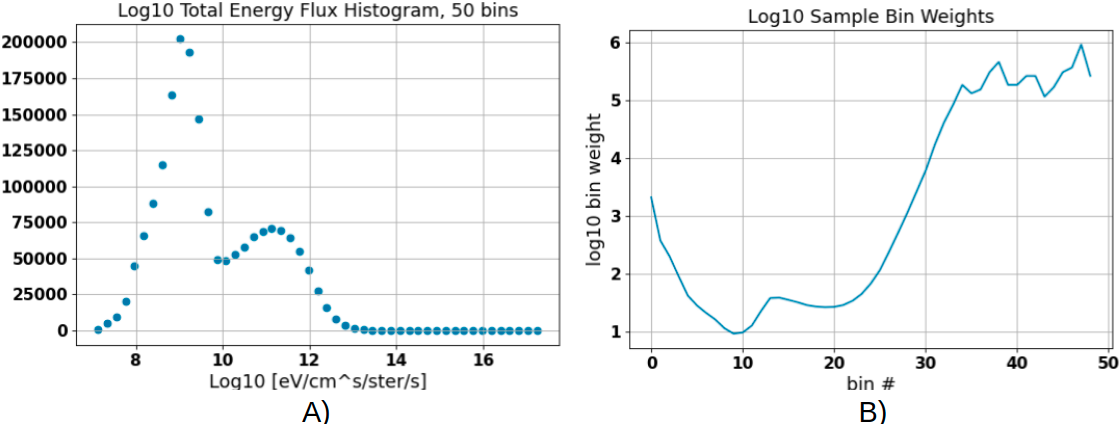}}
\caption{Histogram and weights: Each sample is weighted inversely by the bin's value in which it resides}
\label{hist_50}
\end{figure}

The Tail Loss model required empirically determining the '$a$' and '$y_r$' values. With a different approach we instead use sample weighting based on a distribution of the sample targets. We take the inverse of the frequency of occurrence (in specific bins) of the testing target data to specify the sample weights. Those weights are multiplied by the MSE contribution to the loss: 
\begin{equation}
\begin{aligned}
\textrm{bins} = \min(\log_{10}(y_{\textrm{train}}),...,\max(\log_{10}(y_{\textrm{train}}),\\
\textrm{weight}_{n} = \frac{1}{ [ \textrm{hist}(\textrm{bin}^i_n) + 1 ] \cdot n_{\textrm{total}}} \label{eq3}
\end{aligned}
\end{equation}
A 50-bin histogram used to construct the distribution-based loss is shown in Figure \ref{hist_50}. The Dist Loss approach rebalances the data in a similar fashion to that of classification models.  The analogy is that each bin is viewed as a class even though we are not using categorical loss. This approach is also applicable to any left or right skewed distribution. 

\subsection{Inverse Two-Dimensional Convolutional Model}
All of the models discussed thus far predict one spatial value per one model evaluation. This is advantageous for keeping the model simple and allows for fast training and testing of specific spacial target points.  It trains the model to learn 2D features, but in a way that it is indirect and hidden from the model designer. To create a global map from these models we must evaluate them over all desired MLAT-MLT grid points. There is evidence that a 2D convolutional and/or inverse convolutional approach (a review in McCann et al.\cite{McCann_2017} ) can provide significant improvement regarding revealing 2D structures and allowing more complex features to emerge. However in our case, only a sparse spatial representation of the MLAT-MLT ``image pixels'' are available. To create 2D training samples, we combine all concurrent (up to three in orbit on any given year) DMSP satellite data values at each measurement time step.  Our previous models only use one satellite data point at a time and do not couple the same time step data. The input data time cadence is five min. and the target data are available at one-sec. resolution.Therefore, it is an accurate assumption to assume that all spatial locations of data $\pm$ 2.5 min. are at the "same" five-min. training-sample time step. With these assumptions, and a discretization of the MLAT-MLT space by a 128 $\times$ 128 grid, multiple pixel traces of the satellites' paths create known target data points.

\begin{figure}[htbp]
\centerline{\includegraphics[width=0.6\linewidth]{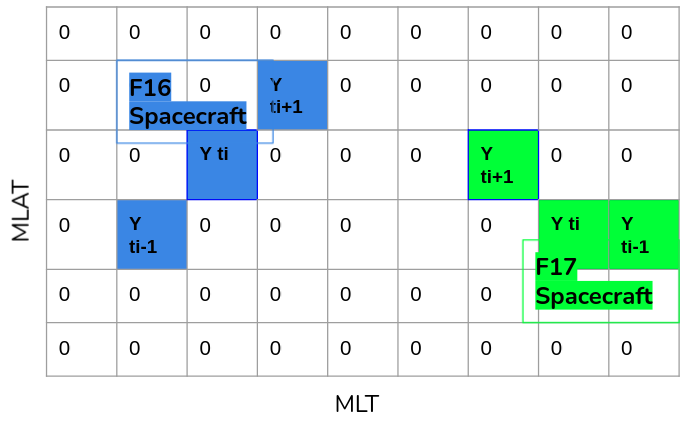}}
\caption{Showing a possible loss function configuration of a 2D training target at a single group of combined timestamps when measured data exists for two DMSP satellites. Only the non-zero values of $y_{\textrm{true}}$ within a $\pm$ 2.5 min. window are used in the loss evaluation.}
\label{2dgrid}
\end{figure}

\begin{figure}[htbp]
\centerline{\includegraphics[width=\linewidth]{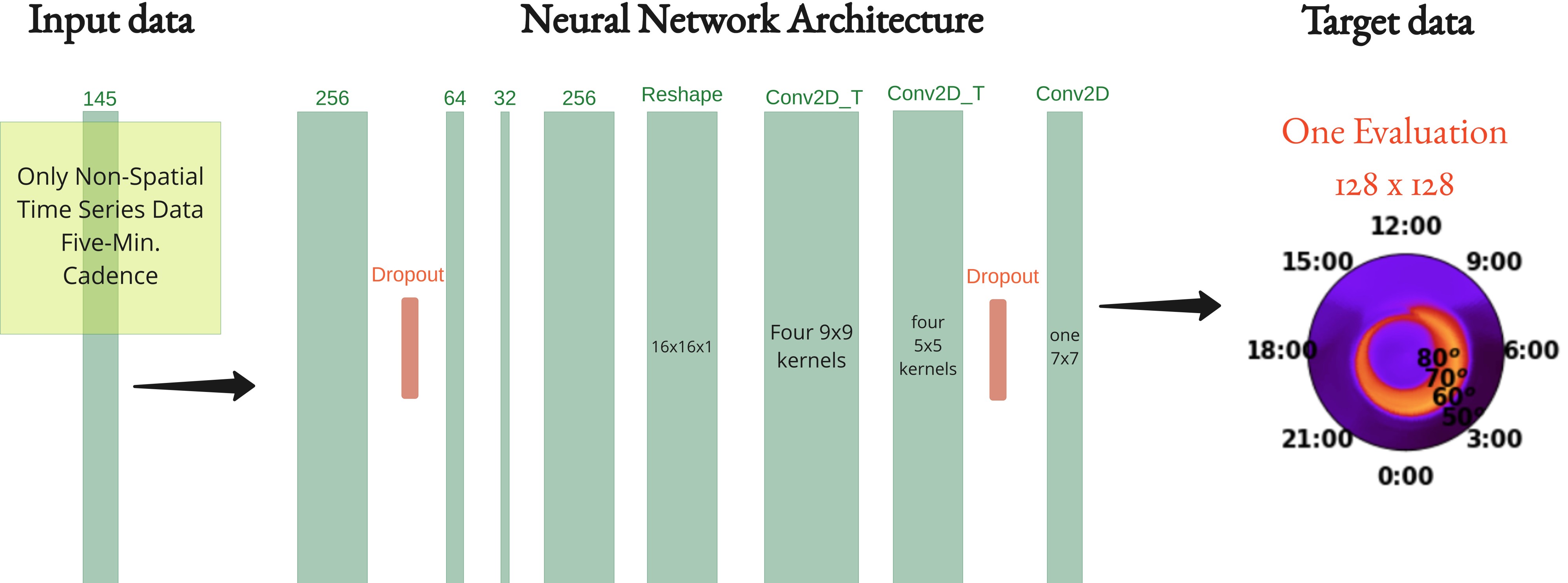}}
\caption{Conv2D Model that takes in no spatial information and predicts the entire northern hemisphere at once.  Spatial information is used to select the sparse cells in the 128x128 grid that are used in the loss function.}
\label{conv2dmodel}
\end{figure}

In the model training we use the custom loss function from Eq \ref{eq4}, where $t_i$ denotes a time range. The square error value of those points for which $y_{true}$ exists in the time range are only used in the summations. The visual in Figure \ref{2dgrid} shows a situation where DMSP F16 and F17 both provide multiple points in colored MLAT-MLT bins
\begin{equation}
\textrm{Loss} =\sum_{t} \sum_{y_{\textrm{true}}(t_i)\exists} ({y_{\textrm{true}}(t_i) - y_{\textrm{pred}}(t_i)})^2 \label{eq4}
\end{equation}
Our network architecture's layers are depicted in Figure \ref{conv2dmodel}, where it is shown that for the first half of the model, the original PrecipNet layers are used up to the 256 node dense layer.  Instead of using the 1024 node layer, two inverse convolutional layers are used (tf.keras.Conv2D\_transpose). These function as normal Conv2D layers, but in reverse, and are therefore also referred to as Con2D\_transpose layers. After these layers a dropout layer and a normal Conv2D follow, where the periodic boundary condition of the MLT grids is applied with an arbitrary 3 cell overlap.  The final Conv2D layer creates the 128x128 sized output.  More that 128x128 resolution is possible, with training time longer than a day.

\section{Model Comparisons}
We evaluate the qualitative characteristics exposed by combining different model architectures and loss functions. 

\subsection{Custom Loss function improvements }
\begin{figure}[htbp]
\centerline{\includegraphics[width=\linewidth]{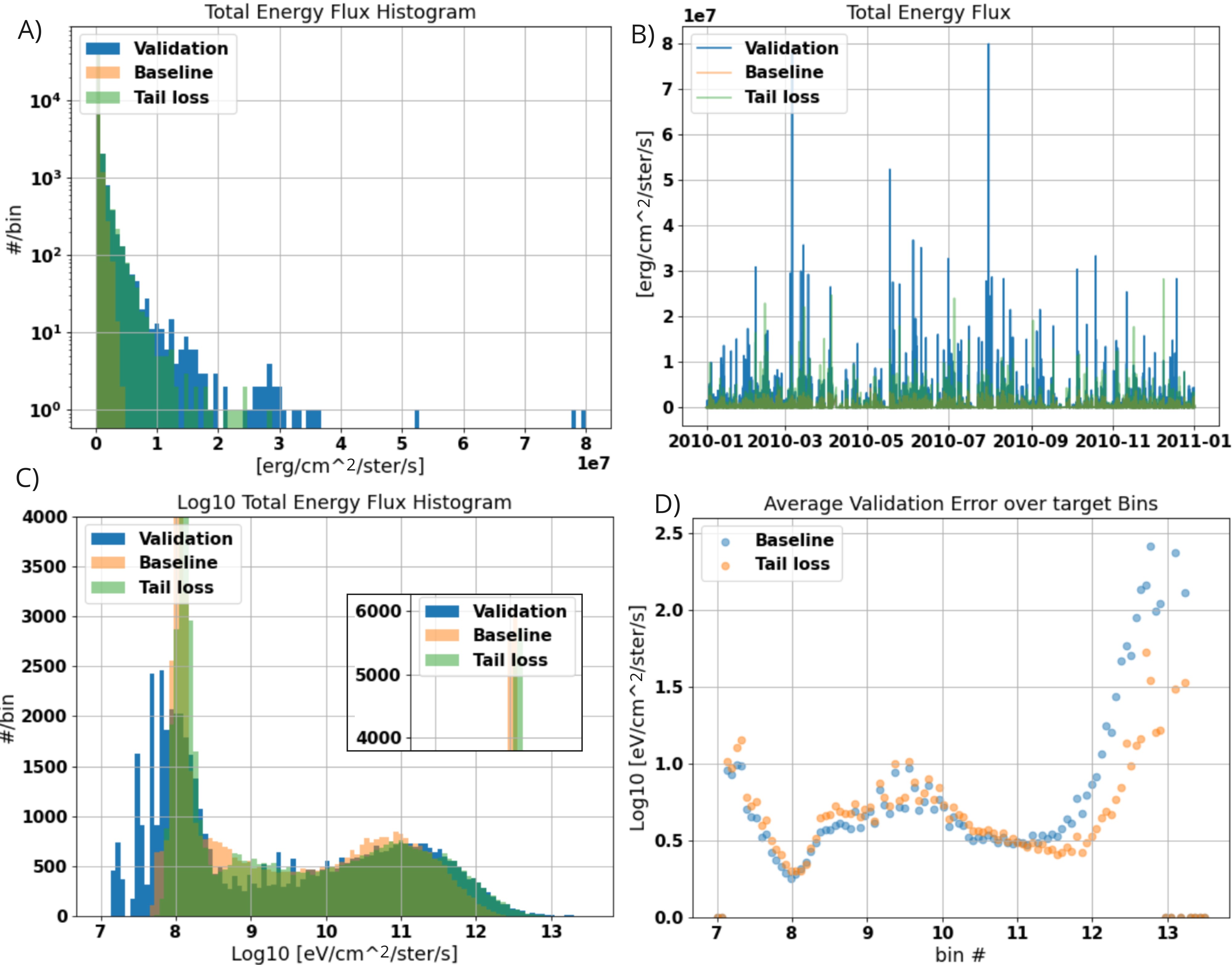}}
\caption{Histograms and errors over "bin-averaged" validation errors.  Notice that after a value of 11 log10 [$eV/{cm}^2/ster/s$] the Tail Loss model has increased accuracy for modeling the physical processes within the the auroral region during space weather storms.}
\label{tail_loss_compare}
\end{figure}

\begin{figure}[htbp]
\centerline{\includegraphics[width=0.82\linewidth]{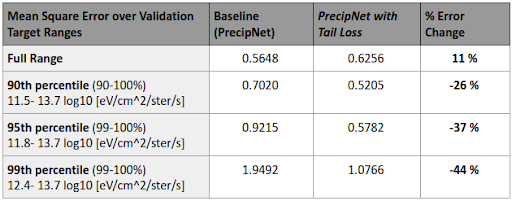}}
\caption{Showing the reduction in error over the right tail region for the 90th, 95th, and 99th percentile, representing the solar storm events to be predicted.}
\label{error_percent_tail_loss}
\end{figure}

MSE regression loss causes large inaccuracies in the high flux tail of the distribution. Therefore, the capability of the Tail Loss model must be measured by improvements in this extreme tail region. In the C) and D) plots of Figure \ref{tail_loss_compare} the Tail Loss function term significantly decreased prediction error for flux values greater than 12 log10 [$eV/{cm}^2/ster/s$], a result of 26\% lower as shown in Table \ref{error_percent_tail_loss}.  Note that transforming out of the log scale, this error difference changes from 250 to 1000\% error. The significance of the accuracy improvement is better visualized by looking at the non-transformed target value histogram (Figure \ref{tail_loss_compare}A  and B), where the model predictions (green) drastically improve the tail specification.  Note that this does not come at the expense of the lower target values at which accuracy is maintained. 
This improvement is even more significant in the 95th and 99th, which represent the most extreme and potentially most impactful space weather storms. This Tail Loss method is a significant improvement to the more commonly known weighted loss which shifts all predictions to the left or right equally.

\subsection{Distribution-based loss improvements}
\begin{figure}[htbp]
\centerline{\includegraphics[width=\linewidth]{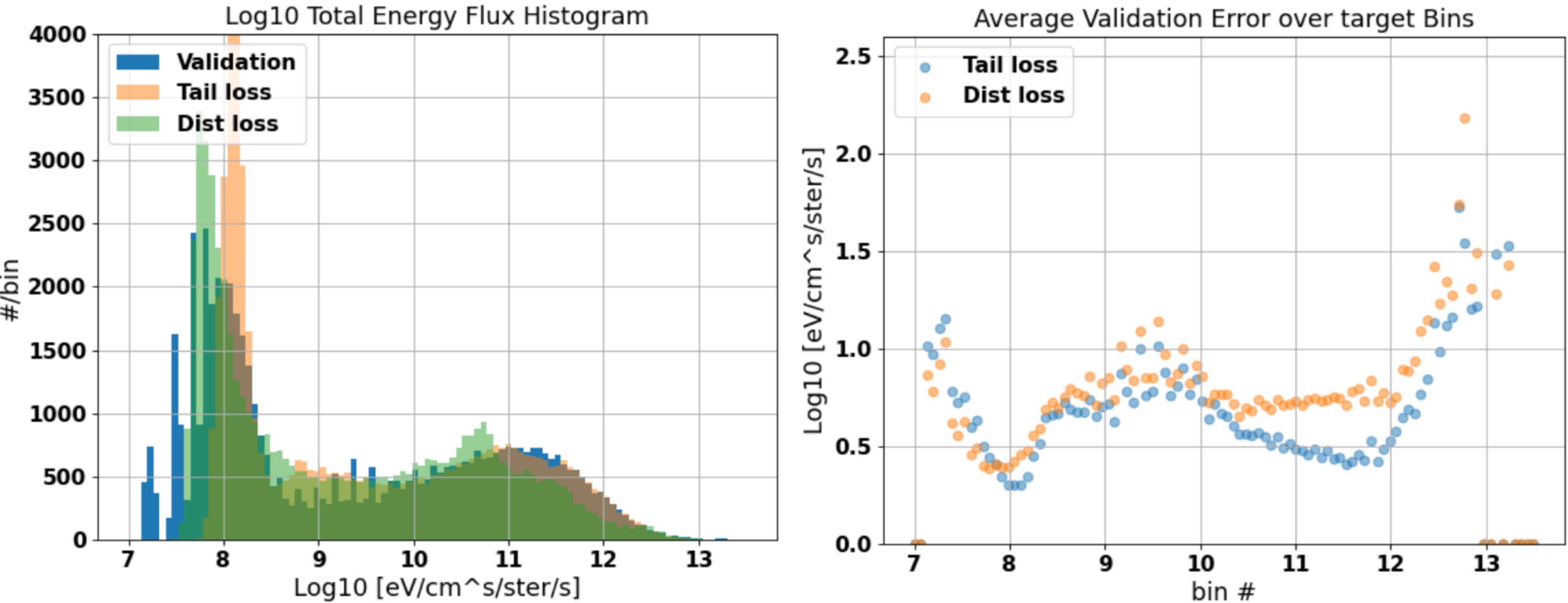}}
\caption{Distribution-loss and tail-loss model histograms and errors.}
\label{dist_loss}
\end{figure}

In Figure \ref{dist_loss}, the distribution loss model is compared to the Tail Loss model. The distribution loss (Dist Loss) considers the imbalance of data across all target data ranges (not just the right-side extreme values), and produces a corresponding reduction in error at both sides of the distribution.   Compared to the PrecipNet baseline model, the right tail distribution and errors are improved by both loss functions, however the Tail Loss model achieves the largest overall accuracy improvement. The Dist Loss results improve over PrecipNet at a log10 target value of 12 and greater, however, the Tail Loss results improve beginning at an order of magnitude lower flux magnitudes (log10 value of 11). It is possible that because the fact that the Dist Loss model is learning both the left and right tails (or more specifically imbalanced data as a whole), the right tail learning is suboptimal. The Dist Loss could be further optimized by customizing the total number of bins used, the applicable bin range for the weights, and the minimum bin size. These optimizations along with the use of non-uniform bin sizes and inclusion of the logic-based approach are the focus of ongoing work.

\subsection{Two-Dimensional Inverse Convolutional Model Results}

\begin{figure}[htbp]
\centerline{\includegraphics[width=0.55\linewidth]{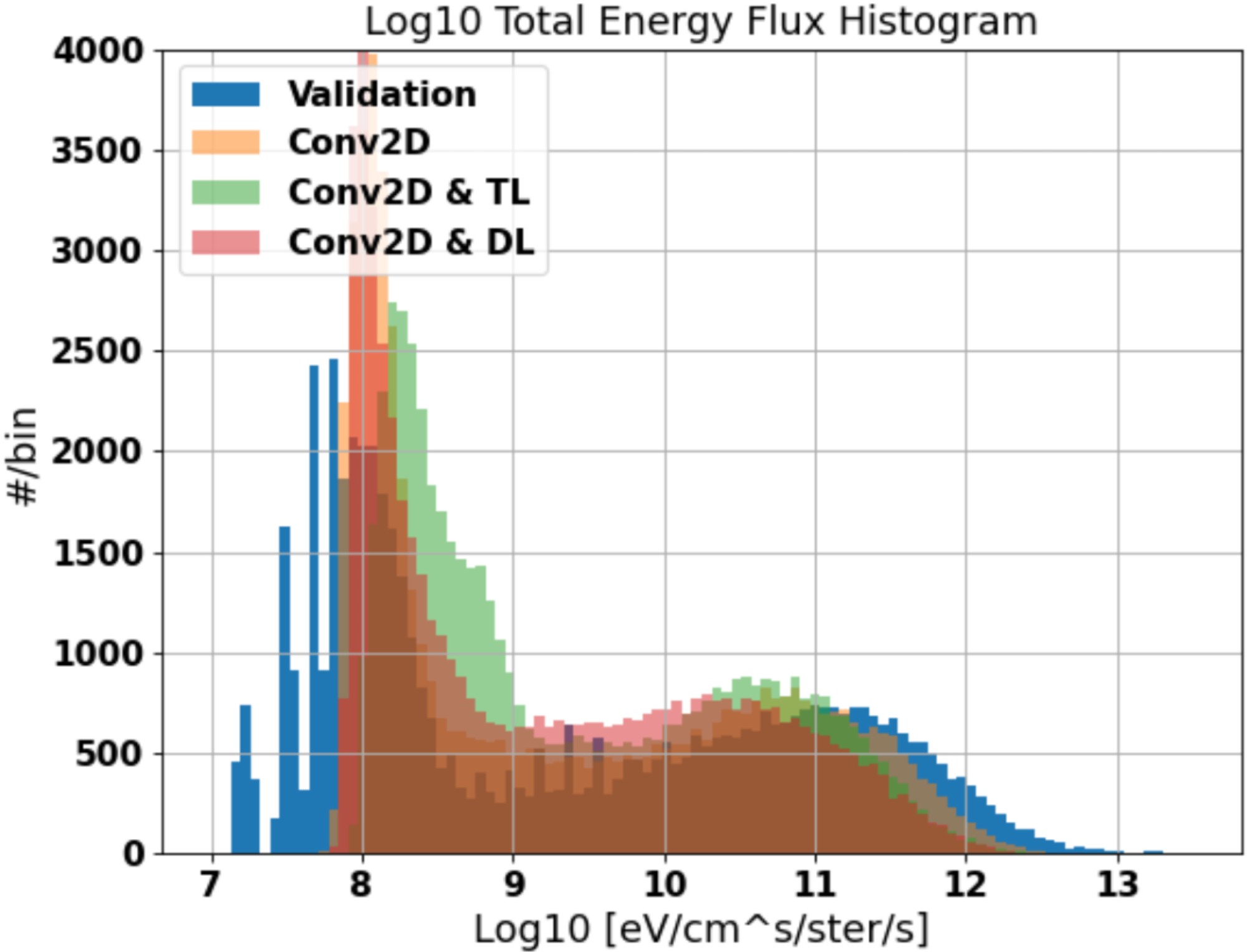}}
\caption{2D inverse Conv2D result histograms.}
\label{tail_loss_2d_hist}
\end{figure}

The Conv2D model has two major potential enhancements: the ability to couple multiple 2D points at one time step and the ability to directly keep track of each training sample on the 2D hemisphere. This model treats the lower and higher MLat values differently because of the ability of the convolutional kernels (the filters) to keep track of the MLat-MLT regions over which they are applied.  The result is a small but noticeable improvement in the upper tail region as well as a lowering and widening of the main left peak as shown in Figure \ref{tail_loss_2d_hist} . We also tested this approach with the Tail Loss and Dist Loss functions, however, Figure \ref{tail_loss_2d_hist} shows that the custom loss functions fail to improve the 2D Conv model.  This may be due to the sensitivity of the discrete kernel-based 2D model to the interaction of weights. For example, the model places emphasis on the spatially-gridded locations of samples, potentially counteracting focus placed on samples from the tail of the distribution.  These results also suggest that the sample weights of the inverse 2D Conv model are also imbalanced spatially with different target ranges, which could be accounted for in the design of the loss function.

\subsection{Auroral Activity Northern Hemisphere Comparisons}
\begin{figure}[htbp]
\centerline{\includegraphics[width=.9\linewidth]{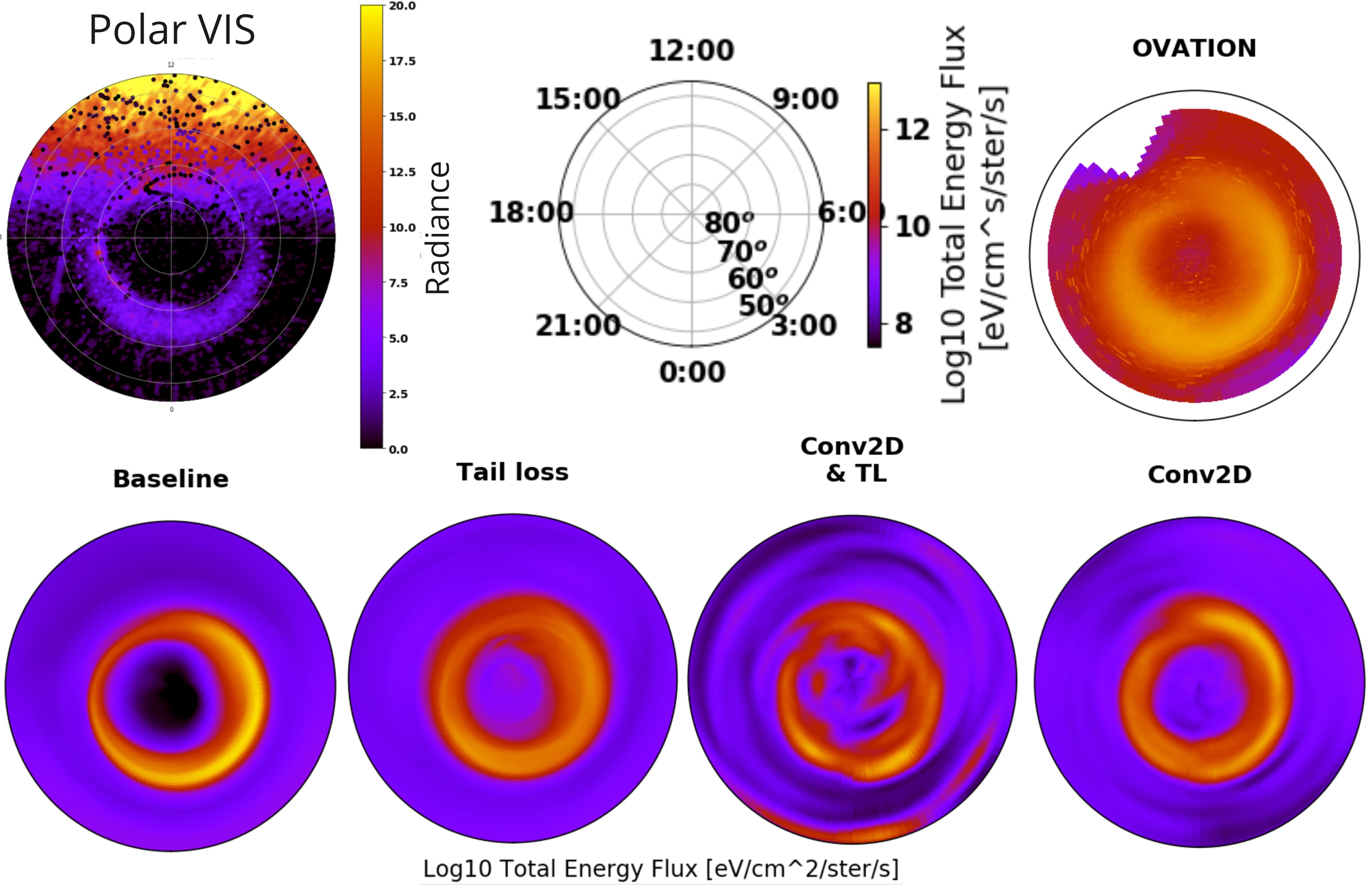}}
\caption{Solar Storm Comparison: Polar VIS experiment results for 1/25/2000 at 18:30. There is no DMSP measurement data at this time step, The upper right plot is blank to serve as visual reference for the labels of the polar plots. The red and yellow colors of the Polar VIS data are from radiance of photons on the day side, not from collisions of charged particles with the atmosphere, and therefore only the black and purple colors are to be compared. }
\label{2000-1-25-18-30_1}
\end{figure}

\begin{figure}[htbp]
\centerline{\includegraphics[width=0.84\linewidth]{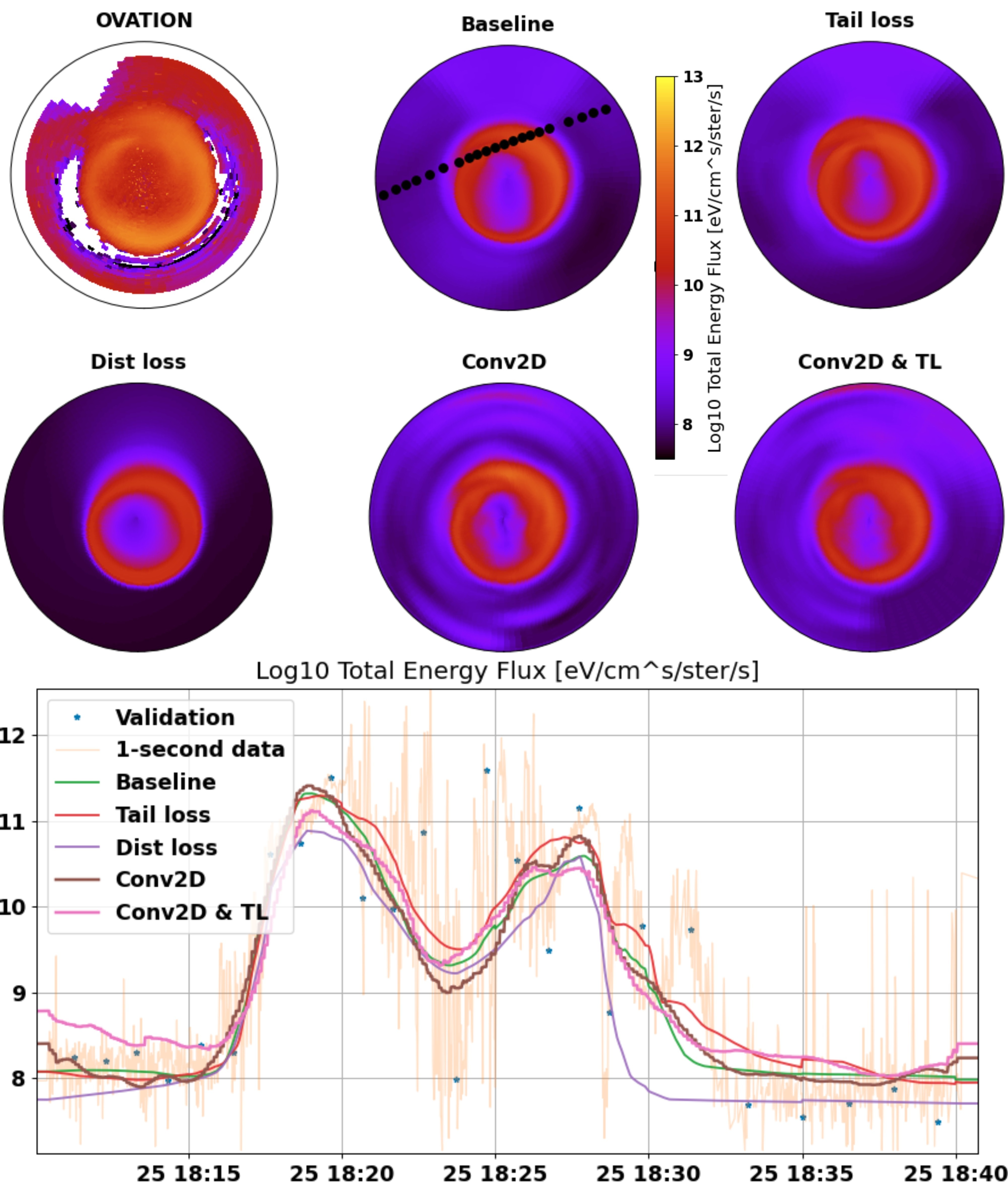}}
\caption{Low activity: 1/25/2010 at 18:30. The validation data are the 60-second cadence subset (black dots on 2D Baseline), used for training, stopping criterion, and MSE error evaluation on 2010 F16 validation data.}
\label{2010-1-25-18-30_1}
\end{figure}

\begin{figure}[htbp]
\centerline{\includegraphics[width=0.84\linewidth]{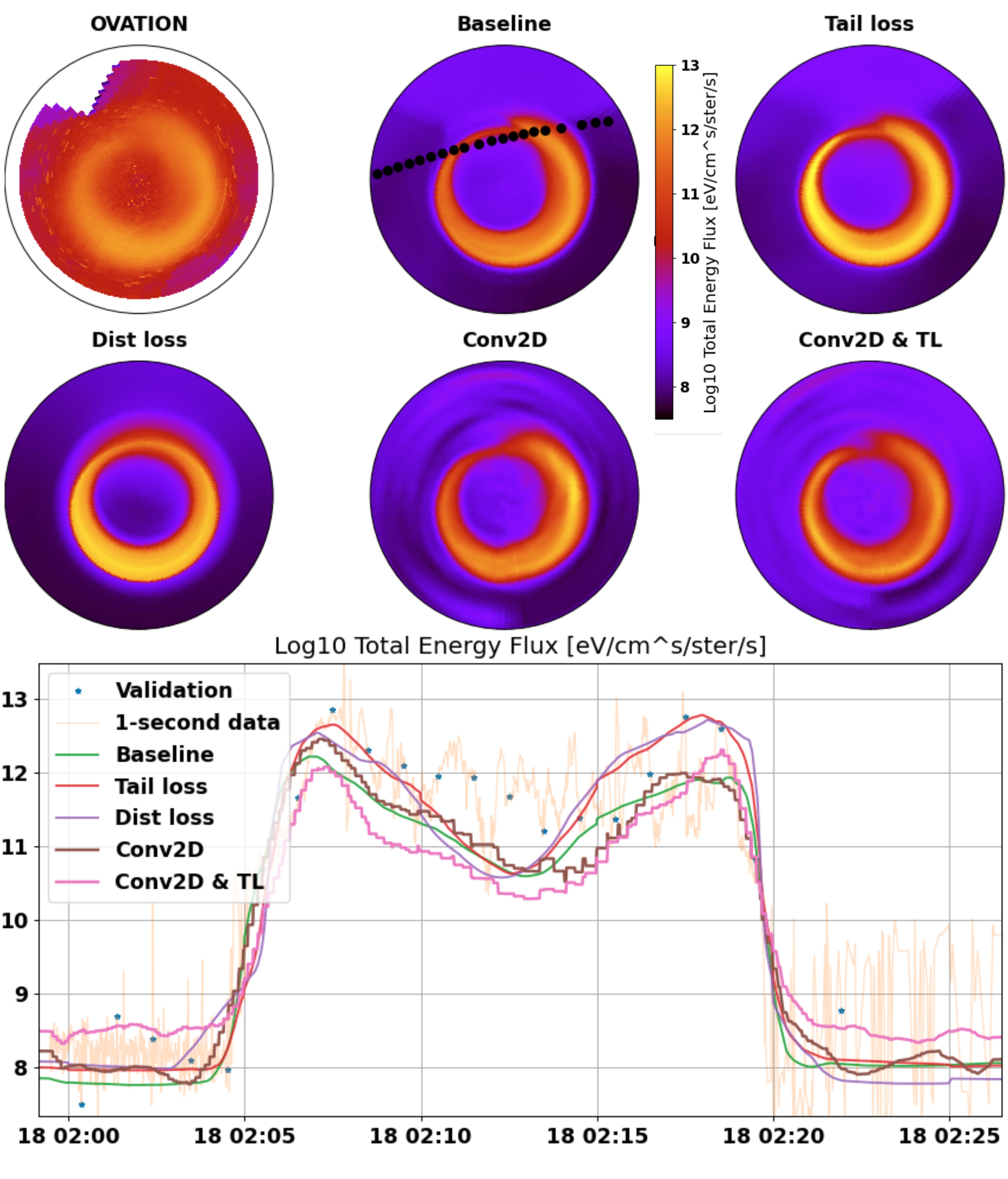}}
\caption{High activity: 3/18/2010 at 2:00. The validation data are the 60-sec. cadence subset (black dots on 2D Baseline), used for training, stopping criterion, and MSE error evaluation on 2010 F16 validation data.}
\label{2010-3-18-2-00_1}
\end{figure}

All models are capable of predicting the full northern hemisphere and this prediction is an important component of evaluation \cite{McGranaghan_2021}. We create these maps for all models presented in this work in Figures \ref{2000-1-25-18-30_1}, \ref{2010-1-25-18-30_1}, and \ref{2010-3-18-2-00_1} for time steps that occur during distinct space weather activity (January 25, 2000 18:30 UT, January 25th, 2010 18:30, and March 18, 2010 at 02:00 UT). Northern hemisphere predictions are shown and compared to imagery data from the Polar spacecraft. The Polar visible imaging assembly (VIS instrument) provides auroral imagery data that are a qualitative validation dataset, of sorts, over which to compare morphologies of the models, shown in Figure \ref{2000-1-25-18-30_1}.  On each of the global map plots we also compare to the `state-of-the-practice' model, OVATION. This, in general, over-predicts the auroral boundaries as well as over-predicting the low range of the fluxes near the sub-auroral regions. Comparing all models, the Tail Loss model without the Conv2D variation performs the best in the auroral region, particularly with obtaining the peak flux predictions in the auroral region while preserving the auroral region shape.  However, the Conv2d models predict finer scale spatial features and qualitative comparison to Polar VIS indicates that these are likely physical. We further investigate their physical validity by plotting time series data available at these times for the events in 2010 (Figures \ref{2010-1-25-18-30_1} and \ref{2010-3-18-2-00_1} bottom plots). These time series plots are from the F16 satellite path at which measurements were made every second.  In Figure \ref{2010-1-25-18-30_1}, there are two flux peaks around (log10) magnitude 11 to 12 and these correspond to passing through each side of the auroral oval.  Our models show less variation than the one-sec. data (demonstrative of "mesoscale" structures) because only 60-second cadence data was used for training and validation.  Also the Conv2D model results for the time series appear to have "steps" and this is due to the fixed spatial resolution of 128x128 for the Conv2D model output.
We cannot yet verify the accuracy of these Conv2D model spatial variations over the full 2D predictions, but the variations are physically and qualitatively realistic. Future work will be to use additional space-based imagery data  (such as additional Polar VIS data and DMSP SSUSI images) to verify the complete maps. 
In Figure \ref{2000-1-25-18-30_1}, a qualitative comparison with Polar VIS is shown, however, because it only shows strong fluxes, it is seen that all models improve in the auroral oval. Even though these Polar VIS qualitative comparisons are possible, there are additional complications to do quantitative comparisons, and indeed direct apples-to-apples quantitative comparisons is not possible. In terms of predicting the finer spatial scale phenomenon, the one-sec. data measurement cadence of DMSP is useful, however, these data has a seemingly random component that is difficult to deconvolve from a 1D spatio-temporal trace into a 2D spatial map.  With the input cadence having a five-min. cadence, spatial variations will be more reasonable to model than temporal ones. Future refinements of our models will continue to focus on the non-random spatial phenomenon shown in the one-sec. traces.

\section{Conclusions}
Increasing our ML model expressive capacity yields improved predictions of extreme events and reveals smaller scale phenomenon. A custom Tail Loss function improves accuracy in the 90th percentile of the right tail of the distribution by 26\% (log10 scale) on average without sacrificing accuracy at lower values of the distribution. Additionally, we lay the foundation for a novel convolutional approach that combines data from multiple simultaneous satellite measurements and utilizes the power of Conv2D kernels. 
Our approach is new and we lay out possibilities for data, loss function, and modeling approaches.  A simple but computationally-expensive approach for improving smaller spatial scale identification is running at higher resolution with the Conv2D model and using one-sec. resolution target data as we are currently training this model with 60-sec. sub-sampled data samples. The Conv2D modelling approach is already showing new smaller-scale spatial predictions, the accuracy of which could improve with more kernels, layers, and grid resolution.

\section*{Acknowledgment}
We thank the \href{https://omniweb.gsfc.nasa.gov/ow_min.html}{NASA OMNI initiative}, and those who provided the \href{https://www.ngdc.noaa.gov/stp/satellite/dmsp/}{Defense Meteorological Satellite Program} (DMSP) data over the several decade lifetime of these spacecraft.

\bibliographystyle{IEEEtran}
\bibliography{IEEEabrv,bibi.bib}
\end{document}